\documentclass[preprint,12pt]{elsarticle}

\usepackage{amssymb}

\usepackage{mathrsfs}
\usepackage{times}
\usepackage{helvet}
\usepackage{courier}
\usepackage{epsfig}
\usepackage{graphicx}
\usepackage{amsmath, amsthm, amssymb}
\usepackage[ruled,linesnumbered]{algorithm2e}
\usepackage{subfigure}

\def\mvp{\vspace*{-0.1in}}

\def\calF{\mathcal{F}}
\def\calH{\mathcal{H}}
\def\calP{\mathcal{P}}
\def\calA{\mathcal{A}}
\def\calT{\mathcal{T}}
\def\calM{\mathcal{M}}
\def\calR{\mathcal{R}}
\def\calX{\mathcal{X}}
\def\calU{\mathcal{U}}

\providecommand{\size}[1]{}
\renewcommand{\size}[1]{{\left|#1\right|}}
\newcommand{\Ignore}[1]{}

\DeclareMathOperator{\sgn}{sgn}
\DeclareMathOperator{\argmax}{argmax}

\allowdisplaybreaks[4]

\journal{Artificial Intelligence}

\begin{document}

\begin{frontmatter}



\title{Learning Probabilistic Hierarchical Task Networks
to Capture User Preferences}


\author{Nan Li}
\ead{nli1@cs.cmu.edu}

\address{Computer Science Department, Carnegie Mellon  University, Pittsburgh, PA 15218 USA}
\author{William Cushing, Subbarao Kambhampati, and Sungwook Yoon
}
\ead{wcushing@asu.edu, rao@asu.edu, Sungwook.Yoon@asu.edu}

\address{Department of Computer Science, Arizona State
  University, Tempe, AZ 85281 USA}

\begin{abstract}
We propose automatically learning probabilistic Hierarchical Task
Networks (pHTNs) in order to capture a user's preferences on plans, by observing
only the user's behavior.  HTNs are a common choice
of representation for a variety of purposes in planning, including
work on learning in planning.  Our contributions are (a) learning
structure and (b) representing preferences.  In contrast, prior work employing HTNs considers learning
method preconditions (instead of structure) and representing domain
physics or search control knowledge (rather than preferences).
Initially we will assume that the observed distribution of plans is an
accurate representation of user preference, and then generalize to the
situation where feasibility constraints frequently prevent the
execution of preferred plans.  In order to learn a distribution on
plans we adapt an Expectation-Maximization (EM)
technique from the discipline of (probabilistic) grammar induction, taking the
perspective of task reductions as productions in a context-free
grammar over primitive actions.  To account for the difference between
the distributions of possible and preferred plans we subsequently modify this core
EM technique, in short, by rescaling its input.

\Ignore{
While much work on learning in planning has focused on learning domain
physics and search control knowledge, little
attention has been paid to learning user preferences on
desirable plans. Hierarchical task networks (HTNs) are known to
effectively encode user prescriptions about what
constitute good plans. However, manual construction is complex and so
costly and error prone. In this paper, we first propose a novel
approach to learning probabilistic hierarchical task networks that
capture user preferences by examining only user-produced plans
(neither the structure of the task network nor the preferences to be
modeled are elicited).  In this algorithm, we assume the
distribution of user-produced plans directly reflects user
preference. We show that this problem
has close parallels to the problem of probabilistic grammar
induction; we adapt grammar induction methods to learn task networks. However, executed
plans are not always an accurate representation of user preferences, as they
result from the interaction between user preferences and feasibility
constraints.  We extend our approach to this situation of preferences
obfuscated by feasibility constraints; we assume that a set of
alternatives to any observed plan is available, and use this
information to rescale the input to the core learning technique in order to better reflect a
user's ideal distribution on plans.  We
will empirically demonstrate the effectiveness of both approaches
separately.}

\end{abstract}





\end{frontmatter}


\section{Introduction}
\begin{figure} 
  \begin{center}
    \includegraphics[width=1\textwidth]{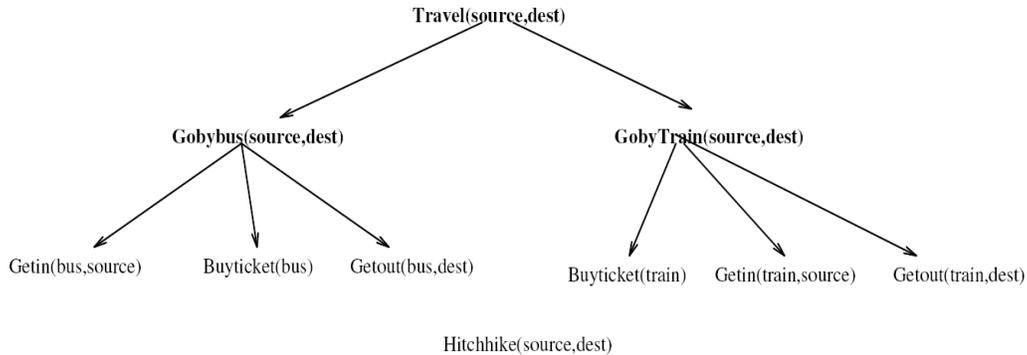}

  \caption{Hierarchical task networks in a travel domain.}
\mvp
  \label{fig:travel}
  \end{center}
\end{figure}

Application of learning techniques to planning is an area of long
standing research interest. Most work in this area to-date has, however,
only considered learning domain physics or search control.  The
relatively neglected alternative, and the focus of this work, is
learning preferences.\Ignore{focused on learning either search control
  knowledge, or domain physics. Another critical piece of knowledge needed for effective plan synthesis
is that of user preferences about desirable plans, and to our knowledge there has not been any
work focused on learning it.} It has long been understood that users
may have complex preferences on plans (c.f. \cite{baier-aimag}). An
effective representation for preferences (among other possible purposes) is a
Hierarchical Task Network (HTN).\Ignore{  Perhaps the most popular 
approach for specifying preferences is by hierarchical task networks
(or HTNs), where} In addition to domain physics (in terms
of primitive actions and their preconditions and effects), the planner
is provided with a set of tasks (non-primitives) and methods (schemas)
for reducing each into a combination of primitives and sub-tasks
(which must then be reduced in turn).\Ignore{   Figure~\ref{fig:travel} shows an
HTN for a travel domain.} A plan (sequence of primitive actions) is considered valid
if and only if it (a) is executable and achieves every specified goal,
and (b) can be produced by recursively reducing a specified task (the
top-level task).  \Ignore{While the
first clause focuses on goal achievement, the second clause ensures
that the plan produced is one that satisfies the user preferences.}

For the example in Figure~\ref{fig:travel} the top level task is to
travel (and the goal is to arrive at some particular destination);
acceptable methods reduce the travel task to either \emph{Gobytrain} or
\emph{Gobybus}.\Ignore{  
goal of traveling from a source to a destination can be achieved
either by {\it Gobytrain}, which involves a specific sequence of
tasks, or {\it Gobybus}.}  In contrast, the plan of hitch-hiking
(modeled as a single action), while executable and goal achieving, is
not considered valid --- the user in question loathes that mode of
travel.  In this way we can separately model physics and (boolean)
preferences; to accommodate degree of preference (i.e., more than just
accept/loathe) we attach probabilities to the methods reducing tasks (and equate
probable with preferred), arriving at probabilistic Hierarchical Task Networks
(pHTNs).

\Ignore{from
the source to the destination, while executable, is not considered a
valid plan. The reduction schemas can be viewed as providing a
``grammar'' of desirable solutions, and the planner's job is to find
executable plans that are also grammatically correct.}

While pHTNs can effectively model preferences, manual
construction (i.e., preference elicitation) is complex, error prone, and costly.  In this paper,
we focus on automatically learning, i.e.\ by observing only user behavior, pHTNs capturing user preferences.\Ignore{learning this grammar,
given only successful plans known to be acceptable to the users.
} Our approach takes off from the view of task networks as
grammars~\cite{geib07,rao98}. First, as mentioned, we generalize by considering pHTNs rather than HTNs (to
accommodate degree of preference).\footnote{Of course pHTNs can be
  trivially converted to HTNs if desired, by simply ignoring the
  learned weights (and if desired, to prevent overfitting perhaps, removing particularly unlikely
  reductions by setting some threshold).}  So each task is associated with a
distribution over its possible reduction schemas, and probable plans
are interpreted as preferred plans. Then we exploit the
connection between task reduction schemas and production rules (in grammars) by adapting the
considerable work on grammar induction
\cite{collins97,charniak00,lari90}.  Specifically, we view plans as
sentences (primitive actions are seen as words) generated by a target grammar, adapt an 
expectation-maximization (EM) algorithm for learning that grammar
(given a set of example plans/sentences), and interpret the result as
a pHTN modeling user preference.

Note that in the foregoing we have assumed that the distribution of example plans directly
reflects user preference.  Certainly preferred plans will be executed
more often than non-preferred plans, but with equal certainty, reality
often forces compromise.  
\Ignore{However, one
challenge here is that executed plans are not an accurate reflection
of the user's true preferences, but rather preferences modulated by
feasibility constraints.} For example, a (poor) graduate student may
very well prefer, in general, to travel by car, but will nonetheless
be far more frequently observed traveling by foot.  In other words, by observing the plans
executed by the user we can relatively easily learn what the user
\emph{usually does},\footnote{A useful piece of knowledge in the
plan recognition scenario~\cite{geib07}.} and so can predict their
behavior as long as feasibility constraints remain the same. It is a
much trickier matter to infer what the user truly \emph{prefers to
do}, and it is this piece of knowledge that would allow predicting
what the user will do in a novel (and improved) situation.
Towards this end, in the second part of the paper, we describe a novel, but intuitive, extension of the
core EM learning technique that rescales the input
distribution in order to undo possible filtering due to feasibility
constraints. The idea is to automatically
generate (presumably less preferred) alternatives (e.g.\ by an automated planner) to the user's
observed behavior and use this additional information to appropriately
reweight the distribution on observed plans.

In the following sections, we start by formally stating the problem of
learning probabilistic hierarchical task networks (pHTNs).  Next, we
discuss the relations between probabilistic grammar induction and pHTN
learning, and present an algorithm that acquires pHTNs from
example plan traces. The algorithm works in two phases. The first
phase hypothesizes a set of schemas that can cover the training
examples, and the second is an expectation maximization phase that
refines the probabilities associated with the schemas.  We then
evaluate our approach against models of users, by comparing the
distributions of observed and predicted plans. Subsequently we
consider possible obfuscation from feasibility constraints and
describe our rescaling technique in detail.  We go on to demonstrate
its effectiveness against randomized models of feasibility constraints.
Finally we discuss related work and summarize our contributions.








\section{Probabilistic Hierarchical Task Networks}

\noindent{\bf{Definitions.}} A pHTN \emph{domain} $\calH$ is a 3-tuple, $\calH = \langle \calA, \calT, \calM \rangle$,
where $\calA$ is a set of primitive actions, $\calT$ is a set of
tasks (non-primitives), and $\calM$ is a set of methods (reduction
schemas).  A pHTN \emph{problem} $\calR$ is a
3-tuple, $R= \langle I,G,T \rangle$, with $I$ the initial state, $G$
the goal, and $T \in \calT$ the top level task to be reduced.
Each \emph{method} $m \in \calM$ is a
$(k+2)$-tuple, $\langle Z, \theta, m_1,m_2,\ldots,m_k \rangle$, where each
$m_i$ is a task or primitive and $\theta$ is the probability of reducing
$Z$ by $m$: let $\calM(Z)$ denote all methods that can reduce $Z$, then
$\sum_{m \in \calM(Z)} \theta(m) = 1$.  Without loss of
generality,\footnote{Any CFG can be put in Chomsky normal form by
  introducing sufficiently many auxiliary non-primitives.  This
  remains true for probabilistic context-free grammars.}  we 
restrict our attention to Chomsky normal
form: each method decomposes a task into either two tasks or one
primitive.  So for any method $m$, either $m = \langle Z, \theta, X,Y
\rangle$ (also written $Z \rightarrow XY,\theta$), with $X,Y \in \calT$, or
$m = \langle Z, \theta, a \rangle$ (also written $Z\rightarrow a,\theta$) with
$a \in \calA$.  Table \ref{travel_htn_2} provides an example of a pHTN
domain in Chomsky normal form modeling the Travel domain (see
Figure~\ref{fig:travel}), in the hypothesis space of our learner (hence
the meaningless task names).  According to the table, the user prefers
traveling by train (80\%) to traveling by bus (20\%).  

For primitives, we follow STRIPS semantics: Each primitive
action defines a transition function on states, and from an initial
state $I$ executing some sequence $a_1,a_2,\ldots,a_k$ of primitives produces a
sequence of states $s_0=I,s_1=a_1(s_0),s_2=a_2(s_1),\ldots,s_k=a_k(s_{k-1})$, provided
each $a_i$ has its preconditions satisfied in $s_{i-1}$.  Such a
sequence is goal-achieving if the goal $G$ is satisfied in the final
state, $s_{k}$ (goals take the same form as preconditions).  

Concerning tasks, a primitive sequence $\phi$ is a \emph{preferred
solution} if there exists a parse of $\phi$ by the methods of $\calH$ with root $T$
(\emph{preferred}), and $\phi$ is executable from $I$ and achieves $G$ (\emph{solution}). A \emph{parse} $\calX$ of $\phi$ by $\calM$
with root $T$ is a tree, more specifically a rooted almost-binary
directed ordered labeled tree, with (a) root labeled by $T$, (b) leaves labeled by $\phi$ (in
order), and (c) each internal vertex is decomposed into its children
(in order) by some $m \in \calM$.  For such internal vertices, say $v$,
let $T(v)$, $m(v)$, and $\theta(v)$ be the associated task, reducing
method, and prior probability of that reduction.  
The prior probability of an entire parse tree is the product of $\theta(v)$ over every internal
vertex $v$.  Given a fixed root, the prior probability of a primitive
sequence is the sum of prior probabilities of every parse of that
sequence with the fixed root: 
\begin{align}
P(\phi \mid \calH,T) &= 
  \sum_{\text{parse } \calX} P(\calX \mid T,\calH)\\
&= 
  \sum_{\text{parse } \calX} \prod_{\;\text{internal } v}  \theta(v);
\end{align}
note that the prior probability of a primitive sequence has nothing to
do with whether or not the sequence is goal-achieving or even
executable.  Enumerating all parses could, however, become expensive, so
in the remainder we approximate by considering only the most probable parse
of $\phi$ --- define:
\begin{align}
\calX^*(\phi) &= \argmax_{\text{parse } \calX \text{ of } \phi}
\prod_{\text{internal } v} \theta(v).
\end{align}

\noindent{\bf Learning pHTNs.} We can now state the pHTN learning problem formally. Fix the total
number of task symbols, $k$, and fix the first task symbol as the top level task $T$.  Given a set $\Phi$\Ignore{$ =
\{ \phi_1 , \phi_2, \ldots, \phi_n \}$} of observed training plans (so each is
executable and (presumably) goal-achieving), find the most likely
pHTN domain, $\calH^*$.  We assume a uniform prior distribution on
domains with $k$ task symbols, so it is equivalent to maximizing the likelihood of the observation:
\begin{align}
\calH^* &= \argmax_\calH P(\calH \mid \Phi,T)\\
  &= \argmax_\calH P(\Phi \mid \calH, T) \cdot\frac{P(\calH \mid T)}{P(\Phi \mid T)}\notag\\
  &= \argmax_\calH P(\Phi \mid \calH, T) 
    & \left(\frac{P(\calH \mid T)}{P(\Phi \mid
  T)}=\frac{\text{uniform prior}}{\text{constant}}\right)\notag\\
  &= \argmax_\calH \prod_{\phi \in \Phi} P(\phi \mid \calH,T).
\end{align}

\Ignore{
However, this definition has (at least) two downsides, one, the computation of
$P(\phi \mid \calH,T)$ could become large --- all parses of $\phi$ could
be a large set --- and two, unambiguous grammars do not enjoy any
advantage over ambiguous grammars.  More specifically, highly probable
plans in an unambiguous grammar are synonymous with highly probable
parse trees, in contrast, highly probably plans in ambiguous grammars
are of two types: those with highly probably parse trees and those
with many improbable parse trees.  We take the perspective that parse
trees amount to justifications for the particular course of 
action taken, and that a single strong justification is superior to many
weak arguments.  One could formalize such a mental process as equations
like $P( \forall \phi' \in F, P( X \mid \phi, \calH, T) > P(X' \mid
\phi', \calH, T) \mid F, \calH, T)$,
where $F$ is the set of alternatives a user is aware of in a
particular situation, but such an approach is very complex.  A similar
effect can be had by just maximizing most probable parses instead:
\begin{align}
\hat{\calH_k} &= \argmax_\calH \prod_{\phi \in \Phi} \max_X P(X \mid
\phi, \calH, T)\\
&= \argmax_\calH \prod_{\phi \in \Phi} \prod_{\;d^+_{X^*}(v)} \theta(v).
\end{align}
}

\noindent{\bf Remark.} The preceding incorporates several simplifying assumptions, most importantly
we are making the connection to context-free grammars as strong as
possible.  In particular our definitions do not permit conditions in
the statement of methods, so preferences such
as ``If in Europe, prefer trains to planes.'' are not directly expressible (and
so not learnable) in general.  Our definitions nominally permit parameterized actions,
tasks, and methods, but, there is no mechanism for conditioning on parameters (e.g., varying
the probability of a reduction based on the value of a parameter), so
it would seem that even indirectly modeling conditional preference is
impossible.  This is both true and false; if one is willing to
entertain somewhat large values of $k$, then the learning problem can
work with a \emph{ground} representation of a parameterized domain,
thereby gaining the ability to learn subtly different --- or wildly
different --- sub-grammars for
distinct groundings of a parameterized task.  Of course, the
difficulty of the learning task depends very strongly on $k$: in the
following we map terms such as ``(buy ?customer ?vendor ?object
?location ?amount ?currency)'' to symbols by truncation (``buy'') rather than grounding
(``buy\_mike\_joe\_bat\_walmart\_3\_dollars'') for just this
reason.  Future work should consider parameters, and
contextual dependencies in general, in greater depth --- perhaps by
taking the perspective of \emph{feature selection} (truncation and
grounding can be seen as extremes of feature
selection)~\cite{intro-feature-selection, liu-feature-selection}.



%
\begin{table}
  \caption{A probabilistic Hierarchical Task Network in Chomsky normal form\Ignore{ for the Travel domain}.}
  \label{travel_htn_2}
  \vskip 3.5pt
{\small
  \hrule
  \begin{tabbing}
    \quad\=\kill
    \>Primitives: Buyticket, Getin, Getout, Hitchhike; \\
    \>Tasks: $\text{Travel}, A_1, A_2, A_3, B_1, B_2$; \\
    \>$\text{Travel} \rightarrow A_2~B_1, 0.2$\qquad\qquad\=$\text{Travel} \rightarrow A_1~B_2, 0.8$ \\
    \>$B_1 \rightarrow A_1~A_3, 1.0$\>$B_2 \rightarrow A_2~A_3, 1.0$ \\
    \>$A_1 \rightarrow \text{Buyticket}, 1.0$\>$A_2 \rightarrow \text{Getin}, 1.0$\qquad\qquad\=$A_3 \rightarrow \text{Getout}, 1.0$
    \end{tabbing}
  \hrule
}
\mvp
\end{table}

\section{Learning pHTNs from User Generated Plans}

\Ignore{It is clear that the pHTN as defined above has strong similarity to
probabilistic context free grammars (PCFG). There is a one-to-one
correspondence between the non-terminals of PCFG and the HTN
non-primitive symbols.
}

Our formalization of Hierarchical Task Networks is isomorphic, not
just analogous, to formal definitions of Context Free Grammars (tasks $\leftrightarrow$ non-primitives,
actions $\leftrightarrow$ words, methods/schemas $\leftrightarrow$
production rules); this comes at a price, but, the advantage is that
grammar induction techniques are more or less directly applicable.
The technique of choice for learning probabilistic grammars, and so the choice we
adapt to learning pHTNs, is Expectation-Maximization~\cite{lari90}.

Despite formal equivalence, casting the problem as learning pHTNs
(rather than pCFGs) does make a difference in what assumptions are
appropriate.\Ignore{ concerning the information that will be available for
learning/analysis, and the sacrifices that can be accepted in pursuit
of finishing learning within a reasonable amount of time. (There is
also a difference when considering generalizing to dependencies on
context, as in Section~\ref{feasibility} concerning pHTNs subjected to
feasibility constraints.)}  For example, we do not allow annotations on the primitives of input sequences giving hints
concerning non-primitives; for language learning it is reasonable to
assume that such annotations are available, because the non-primitives
involved are agreed upon by multiple users (or there is no
communication).  In particular information
sources such as dictionaries and informal grammars can be mined
relatively cheaply.  In the case of preference learning for plans, the
non-primitives of interest are user-specific mental constructs
(preferences), and so it is far less reasonable to assume that
appropriate annotations could be obtained cheaply.  So, unlike
learning pCFGs, our system must invent all of its own non-primitive
symbols without any hints.

Our learner operates in two phases.  First a structure hypothesizer (SH)
invents non-primitive symbols and associated reduction schemas (tasks
and methods), as needed, in a greedy fashion, to cover all the
training examples.  
In the second phase, the probabilities of the reduction schemas are
iteratively improved by an Expectation-Maximization (EM) approach.
The result is a local optima in the space of pHTN domains (instead of
$\calH^* = \argmax_\calH P(\calH \mid \Phi,T)$, the global maximum).


\subsection{Structure Hypothesizer (SH)}

We develop a (greedy) structure hypothesizer (SH) in order to generate
a set of methods that can, at least, parse all plan examples, but more
than that, parse all the plan examples without resorting to various
kinds of trivial grammars (for example, parsing each plan example with
a disjoint set of methods).  The basic idea is to iteratively
factor out frequent common subsequences, in particular frequent common
pairs since we work in Chomsky normal form.  We describe the details
in the following; Algorithm~\ref{alg:sh} summarizes in pseudocode.

\begin{algorithm}[t]
\caption{SH(plan examples $\Phi$) returns pHTN $\calH$}
\label{alg:sh}
\SetKwFunction{ShortestPlan}{shortest-plan}
\SetKwFunction{BestSimpleRecursion}{best-simple-recursion}
\SetKwFunction{MostFrequentPair}{most-frequent-pair}
\SetKwFunction{RewritePlans}{rewrite-plans}
\SetKwFunction{Empty}{empty}
\SetKwFunction{InitializeP}{initialize-probabilities}
\SetKw{Not}{not}
\SetKwData{Left}{left}
\SetKwData{Right}{right}
\SetFuncSty{textit}
\small

$\calH := \{ Z_a \rightarrow a \mid a \in \calA \}$  \tcp*{primitive action schemas}
$\RewritePlans(\Phi, \calH)$;\\

\While{\Not $\Empty(\Phi)$}{
    \uCase{$| \phi := \ShortestPlan(\Phi) | \le 2$}{
          \lIf{$|\phi| = 2$}{$\calH := \calH + (T \rightarrow \phi)$;}\\
          \lElse{\label{SH-relaxed-Chomsky}$\calH := \calH \cup 
                      \{ T \rightarrow \alpha \mid Z \rightarrow \alpha \in \calH \}$}
          \tcp*[f]{$\phi = Z$ for some $Z$}}
    \uCase{$(\langle Z,X,d\rangle := \BestSimpleRecursion(\Phi))$ is good enough}{
          \lIf{$d = \Left$}{$\calH := \calH + (Z \rightarrow Z~X)$;}\\
          \lIf{$d = \Right$}{$\calH := \calH + (Z \rightarrow X~Z)$;}} 
    \uCase{otherwise}{
          $(X,Y) := \MostFrequentPair(\Phi)$;\\
          $\calH := \calH + (Z_{XY} \rightarrow
          X~Y)$;\tcp*[f]{$Z_{XY}$ is a new task}}
    $\RewritePlans(\Phi, \calH)$;\tcp*[f]{Plans rewritten to $T$ are removed}
}

$\InitializeP(\calH)$;\\

\Return{$\calH$}

\end{algorithm}

SH learns reduction schemas in a bottom-up fashion.  It
starts by initializing $\calH$ with a separate reduction
for each primitive (from distinct non-primitives); this is a minor
technical requirement of Chomsky normal form.\footnote{It is not
  necessary to use a \emph{distinct} non-primitive for each reduction to a
  primitive, but it does not really hurt either, as synonymous primitives can be
  identified one level higher up in the grammar at a small cost in
  number of rules.} Then all plan examples are rewritten using this
initial set of rules: so far not much of import has occurred.

Next the algorithm enters its main loop: hypothesizing additional
schemas until all plan examples can be parsed to an instance of the
top level task, $T$.\footnote{The implementation in fact allows the single rule $T
  \rightarrow Z$ instead of the set in line \ref{SH-relaxed-Chomsky},
  but for the sake of notation (elsewhere) we assume a strict
  representation here.}
In short, SH hypothesizes a schema, rewrites the plan examples using the new
schema as much as possible and repeats until done.  At that point
probabilities are initialized randomly, that is, by assigning
uniformly distributed numbers to each schema and normalizing by task
(so that $\sum_{m\in \calM(Z)} \theta(m)$ becomes 1 for each task
$Z$) --- the EM phase is responsible for fitting the probabilities to
the observed distribution of plans.  



In order to hypothesize a schema, SH first searches for evidence of a recursive
schema: subsequences of symbols in the
form $\{ sz, ssz, sssz \}$ or $\{ zs,zss, zsss \}$ (simple repetitions).  Certainly
patterns such as $zababab$ have recursive structure, but these are
identified at a later stage of the iteration.  The frequency of
such simple repetitions in the entire plan set is measured, as is their average
length.  If both meet minimum thresholds, then the
appropriate recursive schema is added to $\calH$.  The thresholds
themselves are functions of the average length and total number of
(rewritten, remaining) plan examples in $\Phi$.

If not (i.e., one or both thresholds are not met), then the frequency count
of every pair of symbols is computed, 
and the maximum pair is added as a reduction from a distinct (i.e.,
new) non-primitive.  In the prior example of a symbol sequence
$zababab$, eventually $ab$ might win the frequency count, and be
replaced with some symbol, say $s$.  After rewriting the example
sequence then becomes $zsss$, lending evidence in future iterations, of the kind SH
recognizes, to the existence of a recursive schema (of the form $z
\rightarrow z s$); if such a recursive schema is
added, then eventually the sequence gets rewritten to just $z$.

\begin{figure*} [t]
  \begin{center}
    \includegraphics[width=1\textwidth]{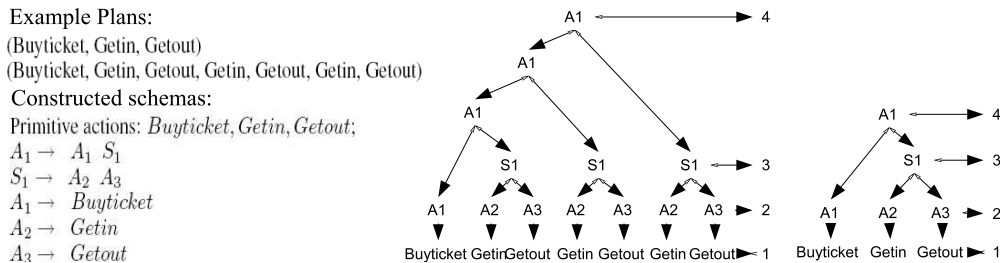}
  \end{center}
  \vskip -20pt
  \caption{A trace of the Structure Hypothesizer on a variant of the Travel domain.}
  \label{fig:vtravel_parse}
\end{figure*}

\medskip
\noindent{\bf Example.}
Consider a variant of the Travel domain (Figure~\ref{fig:travel}) allowing
the traveler to purchase a day pass (instead of a single-trip ticket)
for the train.  Two training plans are shown in
Figure~\ref{fig:vtravel_parse}.  First SH builds the primitive action
schemas: $A_1 \rightarrow \text{Buyticket}$, $A_2 \rightarrow \text{Getin}$, and
$A_3 \rightarrow \text{Getout}$; the updated plan examples 
are shown as level 2 in the Figure.  Next, since $A_2 A_3$ is the most
frequent pair in the plans (and there is insufficiently obvious evidence of
recursion), SH constructs a rule $S_1 \rightarrow A_2 A_3$. After updating the
plans with the new rule, the plans become $A_1 S_1$ and $A_1 S_1  S_1 S_1$, depicted as level 3 in the Figure.  At this point
SH realizes the recursive structure (the simple repetition $A_1 S_1
S_1 S_1$), and so adds the rule $A_1 \rightarrow A_1 S_1$.  After rewriting all plans are parsable to the symbol
$A_1$ (let $T = A_1$), so SH is done: the final set of schemas is at the bottom left
of Figure~\ref{fig:vtravel_parse}.





\subsection{Refining Schema Probabilities: EM Phase}

We take an Expectation-Maximization (EM) approach in order to learn
appropriate parameters for the set of schemas returned by SH.  EM is a
gradient-ascent method with two phases to each iteration: first the current model is used to compute expected values for
the hidden variables (E-step: induces a well-behaved lower bound on the true gradient), and then
the model is updated to maximize the likelihood of those particular
values (M-step: ascends to the maximum of the lower bound).  Doing so will normally change the expected values of the hidden variables, so
the process is repeated until convergence, and convergence does in
fact occur~\cite{probabilistic-graphical-models-book}.  For our
problem, standard (soft-assignment) EM would compute an entire distribution over all
possible parses (of each plan, at each iteration); as the grammars are
automatically generated, there may very well be a huge number of
such parses.  So instead we focus on computing just the parse
considered most likely by the current parameters, that is, we are
employing the hard-assignment variation of EM.  One beneficial
side-effect is that this introduces bias in favor of less ambiguous
grammars --- for in-depth analysis of the tradeoffs involved in choosing
between hard and soft assignment see \cite{kearns-hard-EM,kandylas-wta-EM}.
In the following we describe the details of our specialization of (hard-assignment) EM.

\Ignore{Since all plan examples can be generated by the target reduction
schemas, each plan should have a parse tree associated with it.
However, the tree structures of the example plans $T$ are not
provided. Therefore, we consider $T$ as the hidden variables. We will
use T(o, H) to denote the parse tree of a plan example $o$ given the
reduction schemas $H$. The algorithm operates iteratively. In each
iteration, it involves two steps, an E step, and an M step.}

In the E-step, the current model $\calH_\ell$ is used to compute the most
probable parse tree, $\calX^*_\ell(\phi)$, of each example $\phi$ (from the fixed start
symbol $T$): $\calX^*_\ell(\phi) = \argmax_{\text{parse }\calX}
P(\calX\mid \phi,T,\calH_\ell)$.  This
computation can be implemented reasonably efficiently in a bottom-up fashion
since any subtree of a most probable parse is also a most probable
parse (of the subsequence it covers, given its root, etc.).  The first
level is particularly simple since we associated every primitive, $a$, with
a distinct non-primitive, $Z_a$ (so its most probable, indeed only, parse is just
$Z_a \rightarrow a$).  The remainder of the parsing computes:
\Ignore{
associated with each plan example with symbol $g$ as the root node,
denoted as $ p\>(T\>|\>O,H)$.

To do this, the algorithm computes the most probable parse tree for
each plan example. Any subtree of a most probable parse tree is also
a most probable parse subtree.
Therefore, for each plan example, the
algorithm builds the most probable parse tree in a bottom-up fashion
until reaching the start symbol $g$. For the lowest level, since
each primitive action only associates with one reduction schema of
the form $Z_a \rightarrow a$, the most probable parse trees for them
are directly recorded as their only associated primitive reduction
schemas. For higher levels, the most probable parse tree is decided
by
}
\begin{align}
    P(a_i, &\ldots, a_j \mid Z, \calH_\ell) =\notag\\  
&\max_{k;Z\rightarrow XY,\theta \in \calH_\ell} \theta \cdot 
P(a_i,\ldots,a_k \mid X, \calH_\ell) \cdot 
P(a_{k+1},\ldots,a_j \mid Y, \calH_\ell),
\end{align}
for all indices $i<j \in [n]$ and tasks (non-terminals) $Z$ (so the
parsing computes $O(n^2m)$ maximizations, each in $O(nr/m)$ steps, for a
worst-case runtime of $O(n^3r)$ on a plan of length $n$ with $m$
tasks and $r$ rules in the pHTN).
Conceviably one of the reduction schemas might exist with 0
probability: the implementation prunes such schemas rather than waste
computation.  By recording the rule and midpoint ($k$) winning each
maximization, the most probable parse of $\phi$, $\calX^*_\ell(\phi)$, can be easily extracted
(beginning at $P(a_1,\ldots,a_n \mid T,\calH_\ell)$).

\Ignore{where $o$ is the current action sequence, $a_1, a_2, \ldots a_n$;
$s$ is a reduction schema of the form $a_{root} \rightarrow
a_l\>a_r$, which specifies the reduction schema that is used to
parse $o$ at the first level; $i$ is an integer between $1$ to $n$,
which determines the place that separates $o$ into two subtraces,
$o_1$ and $o_2$. $o_1$ is the action sequence, $a_1, a_2, \ldots
a_i$, and $o_2$ is the action sequence, $a_{i+1}, \ldots a_n$. After
getting s and i, the most probable parse tree of the current trace
consists of $a_{root}$ as the root, and the most probable parse
trees for the subtraces, $T(o_1, H)$ and $T(o_2, H)$, as the left
and right child of the root. The probability of that parse tree is:
$p\>(s\>|\> H) \ast p\>(T(o_1, H)\>|\>o_1, H) \ast  p\>(T(o_2,
H)\>|\>o_2, H)$. This bottom-up process continues until it finds out
the most probable parse tree for the entire plan.}

After getting the most probable parse trees for all plan examples, the
learner moves on to the M-step.  In this step, the probabilities
associated with each reduction schema are updated by maximizing the
likelihood of generating those particular parse trees; let ${\bf X} = \{
\calX^*_\ell(\phi) \mid \phi \in \Phi \}$ and let $M[\text{event}]$ count how many
times the specified event happens in ${\bf X}$ (for example, $M[Z]$, for
some task $Z$, is the total number of times $Z$ appears in the
parses ${\bf X}$).  Then:
\begin{align}
    \calH_{\ell+1} &= \argmax_{\calH'} \prod_{\calX^*\in{\bf X}} P(\calX^*\mid T,\calH'), \notag\\
&= \argmax_{\calH'} \prod_{\calX^*\in{\bf X}} \prod_{\;\;Z\rightarrow XY \in \calX^*} P(Z\rightarrow XY \mid \calH'),\notag\\
&= \argmax_{\calH'} \prod_Z \prod_{\;\;Z\rightarrow
      XY, \theta_{ZXY} \in \calH'} \theta_{ZXY}^{M[Z\rightarrow XY]},
\end{align}
where the maximization is only over different parameterizations (not over
all pHTNs).  Each task $Z$ enjoys independent parameters, and from
above the likelihood expression is a multinomial in those parameters
($\theta_{ZXY} = P(Z \rightarrow XY \mid \calH')$), and so can be maximized simply by
setting:
\begin{align}
P(Z \rightarrow XY \mid \calH_{\ell+1}) &:= \frac{M[Z \rightarrow XY]}{M[Z]}.
\end{align}

That is, the E-step completes the input data $\Phi$ by computing the
parses of $\Phi$ expected by $\calH_\ell$; subsequently the M-step treats
those parses as ground truth, and sets the new reduction probabilities
to the `observed' frequency of such reductions in the completed data.
This improves the likelihood of the model, and the process is repeated
until convergence.

\medskip
\noindent{\bf Discussion:}
Although the EM phase of learning does not introduce new reduction schemas, it does
participate in structure learning in the sense that it effectively
deletes reduction schemas by assigning zero probability to them.  For
this reason SH does not attempt to find a completely minimal
grammar before running EM.  Nonetheless it is important that SH generates
small grammars, as otherwise overfitting could become a serious
problem.  Worst choices of a hypothetical structure learner would include the trivial grammar
that produces all and only the training plans; if this occurs the EM
algorithm above would happily drive the probability of all other rules
to 0 as the included trivial grammar would allow the perfect
reproduction of training data.

\subsection{Evaluation}
To evaluate our pHTN learning approach, we designed
and carried out experiments in both synthetic and benchmark
domains. All the experiments were run on a 2.13 GHz Windows PC with
1.98GB of RAM. Although we focus on accuracy (rather than CPU time),
we should clarify up-front that the runtime for learning is quite
reasonable --- between (almost) 0ms to 44ms per training plan.  We take
an oracle-based experimental strategy, that is, we generate an oracle
pHTN $\calH^*$ (to represent a possible user) and then subsequently use it to
generate behavior $\Phi$ (a set of preferred plans).  Our learner then induces a
pHTN $\calH$ from only $\Phi$; so then we can assess the effectiveness
of the learning in terms of the differences between the original and
learned models.  In some settings (e.g., knowledge discovery) it is
very interesting to directly compare the syntax of learned models against
ground truth, but for our purposes such comparisons are much less
interesting: we can be certain that, syntactically, $\calH$ will look nothing like 
a real user's preferences (as expressed in pHTN form) for the trivial
reason (among others) that $\calH$ will be in Chomsky normal form.  
For our purposes it is enough for $\calH$ to generate an approximately
correct distribution on plans.  So the ideal evaluation is some
measure of the distance between distributions (on plans), for example Kullback-Leibler (KL) divergence:
\begin{align}
D_{KL}( \calP_{\calH^*} \mid\mid \calP_{\calH} ) = &
\sum_\phi \calP_{\calH^*}(\phi)\cdot \log\frac{\calP_{\calH^*}(\phi)}{\calP_{\calH}(\phi)},
\end{align}
where  $\calP_{\calH}$ and $\calP_{\calH^*}$
are the distributions of plans generated by $\calH$ and $\calH^*$
respectively. This measure is 0 for equal distributions, and otherwise
goes to infinity.  

However, as given the summation is over the
infinite set of all plans, so instead we approximate by sampling, but
this exacerbates a deeper problem: the measure is trivially infinite
if $\calP_\calH$ gives 0 probability to any plan (that
$\calP_{\calH_*}$ does not).  So in the following we take measurements
by sampling $X$ plans from $\calH^*$ and $\calH$,
obtaining sample distributions $\hat{\calP}_{\calH^*}$ and
$\hat{\calP}_\calH$, and then we prune any plans not in
$\hat{\calP}_{\calH^*} \cap \hat{\calP}_\calH$, renormalize, obtaining
$\hat{\calP}'_{\calH^*}$ (say $P_1$) and $\hat{\calP}'_\calH$ (say
$P_2$), and finally compute:
\begin{align}
\hat{D}(\calH^* \mid\mid \calH) = D_{KL}( P_1 \mid\mid P_2 ) = &
\sum_\phi P_1(\phi) \cdot \log\frac{P_1(\phi)}{P_2(\phi)}.
\end{align}
This is not a good approach if the intersection is small, but in our
experiments $|\hat{\calP}_{\calH^*} \cap \hat{\calP}_\calH | / |
\hat{\calP}_{\calH^*} \cup \hat{\calP}_\calH |$ is close to 1.  This
modification also imposes an upper bound on the measure, of about
$O(\log X)$.

\subsection{Experiments in Randomly Generated Domains}
\begin{figure*} [t]
 \centering
 \subfigure[]{
   \begin{minipage}[b]{0.48\textwidth}
     \centering \label{fig:learning_rate}
     \includegraphics[width=1\textwidth]{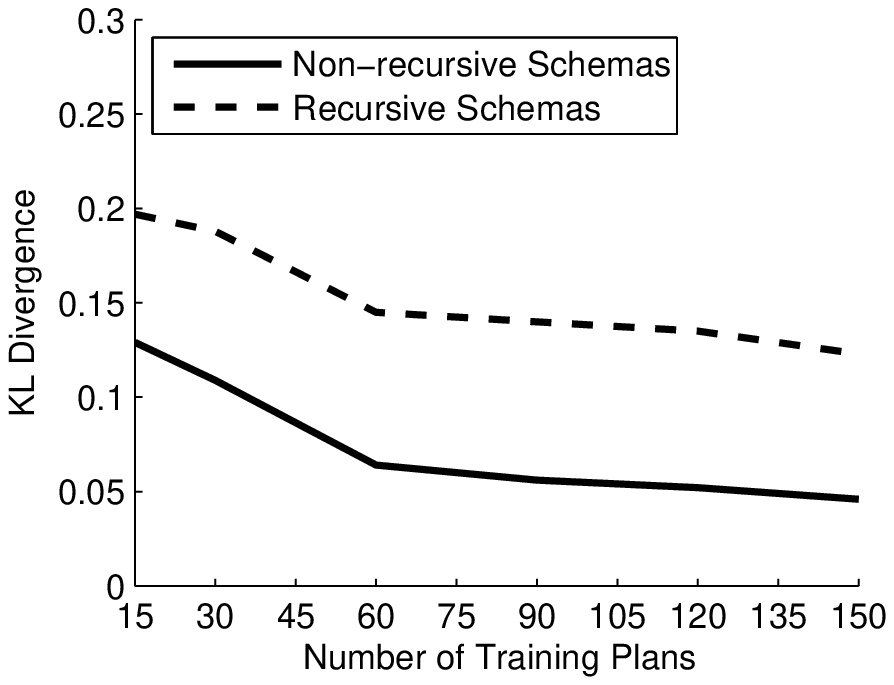}
   \end{minipage}}%
 \subfigure[]{
   \begin{minipage}[b]{0.48\textwidth}
     \centering \label{fig:ratio}
     \includegraphics[width=1\textwidth]{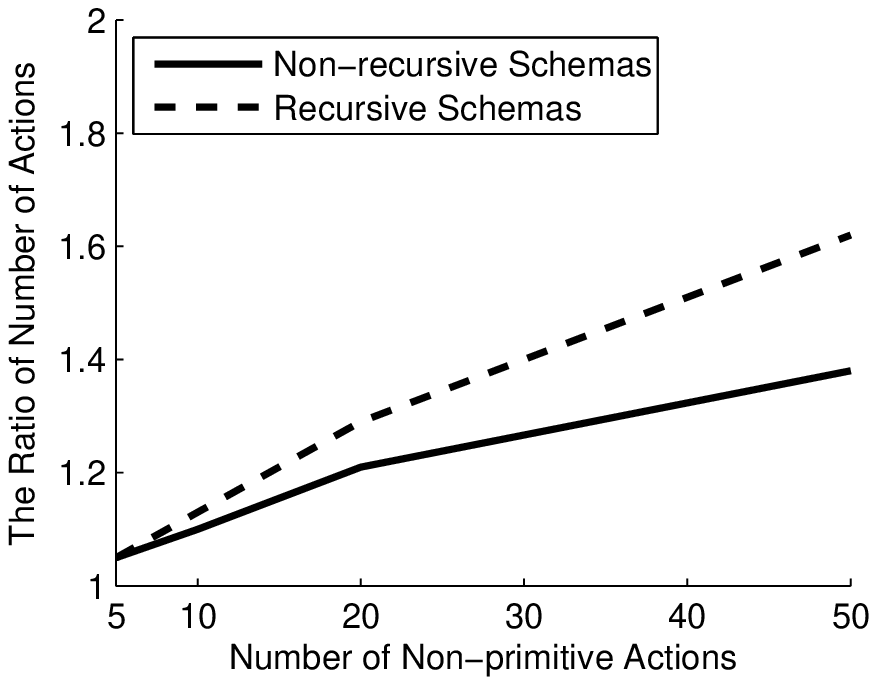}
   \end{minipage}}
 \caption{Experimental results in synthetic domains (a) KL Divergence values with different number of training plans.
 (b) Measuring conciseness in terms of the  ratio between the number
 of actions in the learned and original schemas.}
\mvp
\end{figure*}
In these experiments, we randomly generate the oracle pHTN $\calH^*$,
by randomly generating a set of recursive and
non-recursive schemas on $n$ non-primitives. In non-recursive domains, the randomly
generated schemas form a binary and-or tree with the goal as the
root. The probabilities are also assigned randomly. Generating recursive domains is similar with the only
difference being that  10\% of the schemas generated are recursive.
For measuring overrall performance we provide $10n$ training plans and
take $100n$ samples for testing; for any given $n$ we repeat the
experiment $100$ times and plot the mean.  The results are shown in
Figure~\ref{fig:learning_rate}.  We also discuss two additional, more specialized, evaluations.

\begin{figure*} [t]
 \centering
 \subfigure[]{
   \begin{minipage}[b]{0.48\textwidth}
     \centering \label{fig:kl_nr}
     \includegraphics[width=1\textwidth]{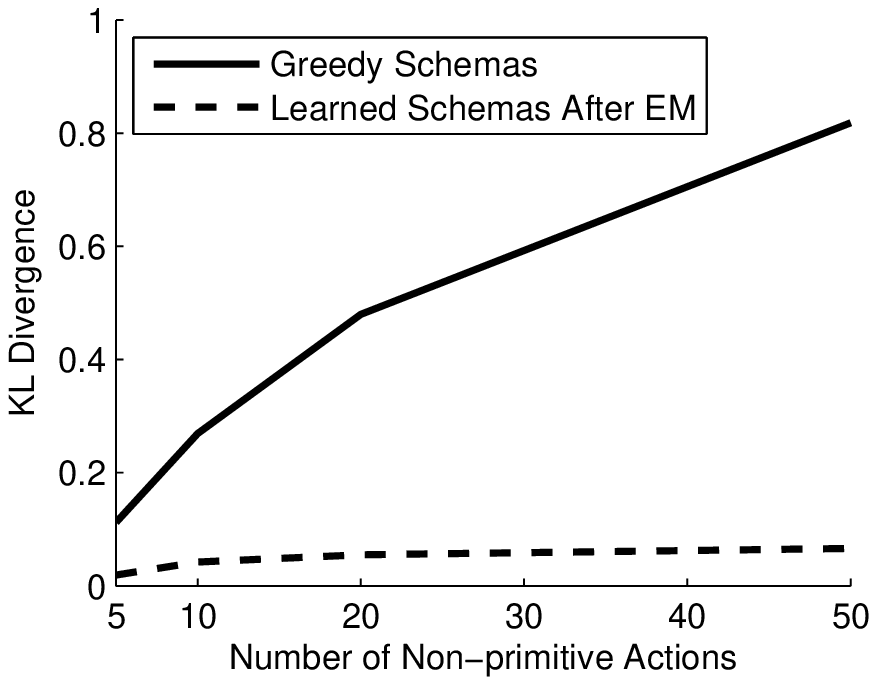}
    \end{minipage}} %
 \subfigure[]{
   \begin{minipage}[b]{0.48\textwidth}
     \centering \label{fig:kl_r}
     \includegraphics[width=1\textwidth]{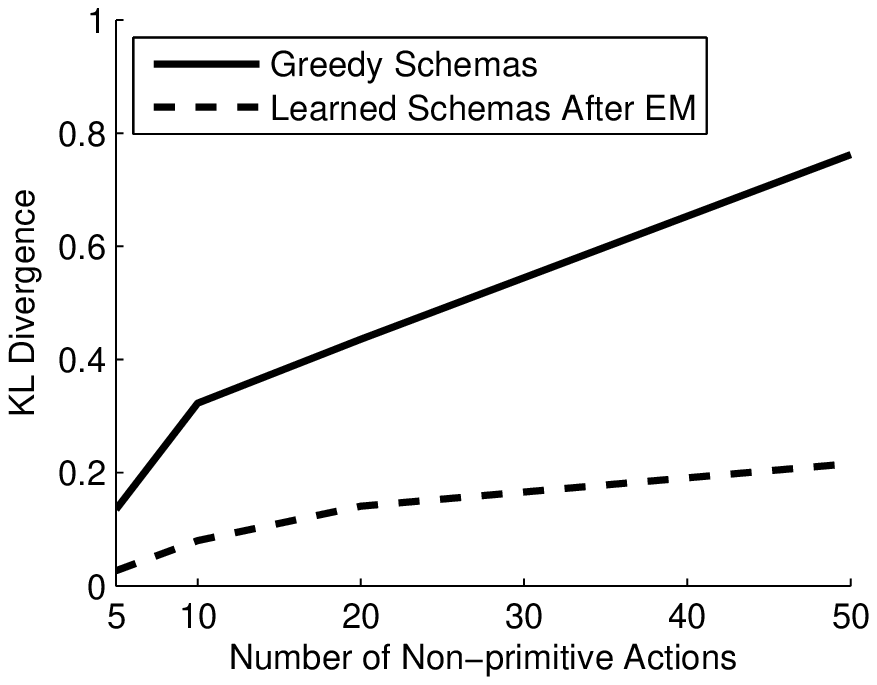}
    \end{minipage}}%
 \caption{Experimental results in synthetic domains  (a) KL Divergence between plans generated by original and learned
 schemas  in {\em non-recursive} domains.
 (b) KL Divergence between plans generated by original and learned
 schemas  in {\em recursive} domains.}
\mvp
\end{figure*}

\medskip\noindent
{\bf Rate of Learning:} In order to test the learning speed, we
first measured KL divergence values with 15 non-primitives
given different numbers of training plans. The results are shown in
Figure~\ref{fig:learning_rate}. We can see that even with a relatively small number of training
examples, our learning mechanism can still construct pHTN schemas
with divergence no more than 0.2; as expected the learning
performance further improves given many training examples.  As briefly
discussed in the setup our measure is not interesting unless the
learned pHTN can reproduce most testing plans with non-zero
probability, since any 0 probability plans are ignored in the
measurement --- so we do not report results given only a very small
number of training examples (the value would be artificially close to
0).  Here `very small' means too small to give at least one example of
every/most reductions in the oracle schema; without at least one
example the structure hypothesizer will (rightly) prevent the
generation of plans with such structure.

\medskip
\noindent {\bf Effectiveness of the EM Phase:} To examine the effect
of the EM phase, we carried out experiments comparing the divergence
(to the oracle) before and after running the EM phase.
Figures~\ref{fig:kl_nr} and~\ref{fig:kl_r} plot results in the non-recursive
and recursive cases respectively.  Overall the EM phase is quite effective,
for example, with 50 non-primitives in the non-recursive setting the EM phase
is able to improve the divergence from 0.818 (the divergence of the
model produced by SH) to the much smaller divergence of 0.066.  

%

%

%

\medskip
\noindent {\bf Conciseness:} The conciseness of
the learned model is also an important factor measuring the quality of the
approach (despite being a syntactic rather than semantic notion),
since allowing huge pHTNs will overfit (with enough available symbols
the learner could, in theory, just memorize the training data). 
A simple measure of conciseness, the one we employ, is the ratio of non-primitives in the
learned model to non-primitives in the oracle ($n$) --- the learner is
not told how many symbols were used to generate the training
data.  Figure~\ref{fig:ratio} plots results.  For small domains (around
$n=10$) the learner uses between
10 and 20\% more non-primitives, a fairly positive result.  However,
for larger domains this result degrades to 60\% more
non-primitives, a somewhat negative result.  Albeit the divergence
measure improves --- on the hidden test set --- so while there is some
evidence of possible overfitting the result is not alarming.  Future
work in structure learning should nonetheless examine this issue
(conciseness and overfitting) in greater depth.

\medskip
\noindent
{\bf Note:}
Divergence in the recursive case is consistently larger
than in the non-recursive case across all experiments:  this is
expected.  In the recursive case the plan space is actually infinite; in the
non-recursive case there are only finitely many plans that can be
generated.  So, for example, in the non-recursive case, it is actually possible for a finite
sample set to perfectly represent the true distribution (simply memorizing the training data will
produce 0 divergence eventually).


\subsection{Benchmark Domains}

In addition to the experiments with synthetic domains, we also
picked two of the well known benchmark planning domains and simulated
possible users (in the form of hand-constructed pHTNs).

\medskip
\noindent{\bf Logistics Planning:} The domain we used in the first
experiment is a variant of the Logistics Planning domain, inside
which both planes and trucks are available to move packages, and every
location is reachable from every other.  There
are 4 primitives in the domain: {\it load},
{\it fly}, {\it drive} and {\it unload}; we use 11
tasks to express, in the form of an oracle pHTN $\calH^*$ (in Chomsky
normal form, hence 11 tasks), our preferences concerning logistics
planning.  We presented 100 training plans to the learning system;
these demonstrate our preference for moving packages by planes rather than
trucks and for using overall fewer vehicles.

The divergence of the learned model is 0.04 (against a hidden test
set, on a single run).  While we are generally unconcerned with the
syntax of the learned model, it is interesting to consider in this
case: Table~\ref{tbl:log} shows the learned model.  With some effort
one can verify that the learned schemas do capture our preferences:  the
second and third schemas for `movePackage' encode delivering a
package by truck and by plane respectively (and delivering by plane
has significantly higher probability), and the first schema permits
repeatedly moving packages, but with relatively low probability.  That
is, it is possible to recursively expand `movePackage' so that one
package ends up transferring vehicles, but, the plan that uses only
one instance of the first schema per package is significantly more
probable (by $0.17^{-k}$ for $k$ transfers between vehicles).


\begin{table}[t]
  \caption{Learned schemas in Logistics}
  \label{tbl:log}
  \vskip 3.5pt
{\small
  \hrule
  \begin{tabbing}
    \quad\=\kill
    \>Primitives: load, fly, drive, unload; \\
    \>Tasks: movePackage, $S_0, S_1, S_2,S_3, S_4, S_5$; \\
    \>movePackage $\rightarrow$ movePackage movePackage, $0.17$ \\
    \>movePackage $\rightarrow S_0~S_5, 0.25$\qquad\qquad\=$S_5 \rightarrow S_3~S_2, 1.0$ \\
    \>movePackage $\rightarrow S_0~S_4, 0.58$\>$S_4 \rightarrow S_1~S_2, 1.0$ \\
    \>$S_0 \rightarrow$ load, $1.0$\>$S_1 \rightarrow$ fly, $1.0$ \\
    \>$S_2 \rightarrow$ unload, $1.0$\>$S_3 \rightarrow$ drive, $1.0$
    \end{tabbing}
  \hrule
}
\mvp
\end{table}

\medskip
\noindent{\bf Gold Miner:}
The second domain we used is Gold Miner, introduced in the learning track of the 2008 International Planning
Competition.  The setup is a (futuristic) robot tasked with
retrieving gold (blocked by rocks) within a mine; the robot can employ
bombs and/or a laser cannon.  The laser cannon can destroy both hard
and soft rocks, while bombs only destroy soft rocks, however, the
laser cannon will also destroy any gold immediately behind its
target.  The desired strategy, which we encode in pHTN form using 12
tasks ($\calH^*$), for this domain is: {\em  1) get the laser cannon, 2) shoot the
  rock until reaching the cell next to the gold, 3) get a bomb, 4) use
  the bomb to get gold.} 

We gave the system 100 training plans of various lengths (generated by
$\calH^*$); the learner achieved a divergence of 0.52.  This is a
significantly larger divergence than in the case of Logistics above,
which can be explained by the significantly greater use of recursion
(one can think of Logistics as less recursive and Gold Miner as more
recursive, and as noted in the random experiments, recursive domains
are much more challenging).  Nonetheless the learner did succeed in
qualitatively capturing our preferences, which can be seen by
inspection of the learned model in Table \ref{tbl:gold}.
Specifically, the learned model only permits plans in the order given
above: get the laser cannon, shoot, get and then use the bomb, and
finally get the gold.




\begin{table}[t]
  \caption{Learned schemas in Gold Miner}
  \label{tbl:gold}
{\small
  \vskip 3.5pt
  \hrule
  \begin{tabbing}
    \quad\=\kill
    \>Primitives: move, getLaserGun, shoot, getBomb, getGold; \\
    \>Tasks: goal, $S_0, S_1, S_2,S_3, S_4, S_5, S_6$; \\
    \>goal $\rightarrow S_0$ goal, $0.78$\qquad\qquad\qquad\=goal $\rightarrow S_1~S_6, 0.22$ \\
    \>$S_0 \rightarrow$ move, $1.0$\>$S_5 \rightarrow S_2~S_0, 1.0$ \\
    \>$S_1 \rightarrow$ getLaserGun, $0.22$\>$S_1 \rightarrow S_1~S_5, 0.78$ \\
    \>$S_2 \rightarrow$ shoot, $1.0$\>$S_6 \rightarrow S_3~S_4, 1.0$ \\
    \>$S_3 \rightarrow$ getBomb, $0.29$\>$S_3 \rightarrow S_3~S_0, 0.71$\\
    \>$S_4 \rightarrow$ getGold, $1.0$
    \end{tabbing}
  \hrule
}
\mvp
\end{table}

\section{Preferences Constrained by Feasibility}
\Ignore{Learning pHTNs from User Generated Plans Obfuscated by
  Feasibility Constraints}
\Ignore{\section{Learning User Preferences from Plans Obfuscated by Feasibility Constraints}}

In general, users will not be so all-powerful that behavior and desire
coincide.  Instead a user must settle for one (presumably the most
desirable) of the feasible possibilities.  Supposing those
possibilities remain constant then there is little point in
distinguishing desire and behavior; indeed, the philosophy of behaviorism \emph{defines}
preference by considering such controlled experiments.  Supposing
instead that feasible possibilties vary over time, then the
distinction becomes very important.  One example we have already
considered is that of a poor grad student: in the rare
situation that such a student's travel is funded, then it would be
desirable to realize the preference for planes over cars.  In addition
to that example, consider the requirement to go to work on weekdays (so the constraint
does not hold on weekends).  Clearly the weekend activities are the
preferred activities.  However, the learning approach developed so far
would be biased --- by a factor of $\frac{5}{2}$ --- in favor of the
weekday activities.  In the following we consider how to account for
this effect: the effect of feasibility constraints upon learning preferences.

\Ignore{As mentioned earlier, although learning preferences by observing user
behavior is a very promising direction, executed plans
may not represent true user preferences due to feasibility
constraints. Therefore, we extend the original approach to uncover the
real preferences embedded in plans produced subject to feasibility
constraints.}

Recall that we assume that we can directly observe a user's behavior, for
example by building upon the work in \emph{plan recognition}.  In this
section we additionally assume that we have access to the set of
feasible alternatives to the observed behavior --- for example by
assuming access to the planning problem the user faced and building upon the work in \emph{automated planning}~\cite{automated-planning-book}.\footnote{Since we
  already assume plan recognition, it is not a signficant stretch to
  assume knowledge of the planning problem itself.  Indeed, planning
  problems are often recast as a broken plan on two dummy actions
  (initial and terminal), and
  solutions as insertions that fix the problem-as-plan.  In particular
  recognizing plans subsumes, in general, recognizing problems.}  For our
purposes it is not a large sacrifice to exclude any number of feasible alternatives
that have never been chosen by the user (in some other situation), in
particular we are not concerned about the potentially enormous
number of feasible alternatives (because we can restrict our attention
to a subset on the order of the number of observed plans).\footnote{Work in \emph{diverse
    planning}~\cite{srivastava07,tuan-diverse} is, then, quite relevant (to picking a subset of
  feasible alternatives of manageable size).} So, in this section, the input to
the learning problem becomes:

{\bf Input.}~~~The $i^\text{th}$ observation, $(\phi_i,F_i) \in \Phi$,
consists of a set of feasible possibilities, $F_i$, along with the
chosen solution: $\phi_i \in F_i$. 

In the rest of the section we consider how to exploit this additional
training information (and how to appropriately define the new learning
task).  The main idea is to rescale the input (i.e., attach weights to
the observed plans $\phi_i$) so that rare situations are not penalized
with respect to common situations.  We approach this from the
perspective that preferences should be transitively closed, and as a
side-effect we might end up with disjoint sets of incomparable plans.
Subsequently we apply the base learner to each, rescaled,
component of comparable plans to obtain a set of pHTNs capturing user
preferences.  Note that the result of the system can now be `unknown'
in response to `is A preferred to B?', in the case that A and B are not
simultaneously parsable by any of the learned pHTNs.  This additional
capability somewhat complicates evaluation (as the base system can
only answer 'yes' or 'no' to such queries).

\subsection{Analysis}

Previously we assumed the training data (observed plans) $\Phi$ was sampled (i.i.d.) directly
from the user's true preference distribution (say $\calU$):
\begin{align*}
P(\Phi \mid \calU) &= \prod_{\phi \in \Phi} P(\phi \mid \calU).
\end{align*}
But now we assume that varying feasibility constraints intervene.  For
  the sake of notation, imagine that such variation is in
  the form of a distribution, say $\calF$, over planning
  problems (but all that is actually required is that the variation is
  independent of preferences, as assumed below).  Note that a planning
  problem is logically equivalent to its solution set.  Then we can write
  $P(F \mid \calF$) to denote the prior probability of any particular
  set of solutions $F$. Since the user chooses
  among such solutions, we have that chosen plans are sampled from the
  posterior, over solutions, of the preference distribution:
\begin{align*}
P(\Phi \mid \calU, \calF) &= \prod_{(\phi, F) \in \Phi} P(\phi \mid
\calU, F) \cdot P(F \mid \calF).
\end{align*}
We assume that preferences and feasibility constraints are mutually
independent: what is possible does not depend upon desire, and desire
does not depend upon what is possible. One can certainly
imagine either dependence --- respectively Murphy's Law (or its
complement) and the fox in Aesop's
fable of Sour Grapes (or envy) --- but it seems to us more reasonable to
assume independence.  Then we can rewrite the posterior of the
preference distribution:
\begin{align*}
P(\Phi \mid \calU, \calF) &= \prod_{(\phi, F) \in \Phi} \frac{P(\phi \mid \calU)}{\sum_{\phi'
    \in F} P(\phi' \mid \calU)} \cdot P(F \mid \calF) &\text{(by assumption).}
\end{align*}
Anyways assuming independence is important, because it makes the
preference learning problem attackable.  In particular, the posteriors
preserve relative preferences\Ignore{ between solutions} --- for all $\phi, \phi' \in F$,
the \emph{odds} of selecting $\phi$ over $\phi'$ are:
\begin{align*}
O(\phi,\phi') &:= \frac{P(\phi \mid \calU, F)}{P(\phi' \mid
 \calU,F)},\\
 &= \frac{P(\phi \mid \calU)}{\sum_{\phi'' \in F} P(\phi'' \mid \calU)} \div
\frac{P(\phi' \mid \calU)}{\sum_{\phi'' \in F} P(\phi'' \mid \calU)},\\
&=\frac{P(\phi \mid \calU)}{P(\phi' \mid \calU)}.
\end{align*}
Therefore we can, given sufficiently many of the posteriors, reconstruct the prior by transitive closure; consider
$\phi,\phi',\phi''$ with $\phi, \phi' \in F$ and $\phi',\phi'' \in F'$:
\begin{align*}
O(\phi,\phi'') 
&= O(\phi,\phi') \cdot O(\phi',\phi''),\notag\\
&= \frac{P(\phi \mid \calU, F)}{P(\phi' \mid \calU, F)} \cdot
\frac{P(\phi' \mid \calU, F')}{P(\phi'' \mid \calU, F')}.
\end{align*}
So then the prior can be had by normalization:
\begin{align*}
P(\phi \mid \calU) = \frac{1}{1 + \sum_{\phi' \neq \phi} O(\phi',\phi)}.
\end{align*}
Of course none of the above distributions are accessible; the learning
problem is only given $\Phi$. Let $M_F[\phi] = \size{\{i \mid (\phi,F) = (\phi_i,F_i) \in \Phi\}}$,
$M_F = \sum_\phi M_F[\phi]$, and $M = \sum_F M_F = \size{\Phi}$.  Then $\Phi$ defines a sampling
distribution:
\begin{align}
\hat{P}(\phi,F \mid \Phi) &= \frac{M_F[\phi]}{M},&\text{so,}\notag\\
\hat{P}(\phi \mid F, \Phi) &= \frac{M_F[\phi]}{M_F},\notag\\
&\approx P(\phi \mid \calU, F) &\text{(in the limit),}\notag\\
\intertext{in particular:}
\hat{O}_F(\phi,\phi') := \frac{M_F[\phi]}{M_F[\phi']} &\approx
\frac{P(\phi \mid \calU)}{P(\phi' \mid \calU)} &\text{(in the
  limit)}
\end{align}
for any $F$ --- but for anything less than an enormous amount of data
one expects $\hat{O}_F$ and $\hat{O}_{F'}$ to differ considerably for
$F \neq F'$, therein lying one of the subtle aspects of the following
rescaling algorithm.  The intuition is, however, simple enough: pick
some base plan $\phi$ and set its weight to an appropriately large
value $w$, and then set every other weight, for example that of
$\phi'$, to $w \cdot \hat{O}(\phi',\phi)$ (where $\hat{O}$ is some
kind of aggregation of the differing estimates $\hat{O}_F$); finally give the weighted set
of observed plans to the base learner.  From the preceding analysis,
in the limit of infinite data, this setup will learn the (closest
approximation, within the base learner's hypothesis space, to the) prior
distribution on preferences.

To address the issue that different situations will give different
estimates (due to sampling error) for the relative preference of one plan to another
($\hat{O}_{F}$ and $\hat{O}_{F'}$ will differ)
we employ a merging process on such overlapping situations.
Consider two weighted sets of plans, $c$ and $d$, and
interpret\footnote{The scaling calculations could produce non-integer
  weights.} the weight of a plan as the number of times it `occurs',
e.g., $w_c(\phi)=M_c[\phi]$. In the simple case that there is only a single plan in
the intersection, $\{\alpha\} = c \cap d$, there is only one way to
take a transitive closure --- for all $\phi$ in
$c$ and $\phi' \in d \setminus c$:
\begin{align*}
\hat{O}_{cd}(\phi,\phi') &= \hat{O}_c(\phi,\alpha) \cdot \hat{O}_d(\alpha,\phi'),\\
  &= \frac{M_c[\phi]}{M_c[\alpha]} \cdot \frac{M_d[\alpha]}{M_d[\phi']},\\
  &= \frac{M_c[\phi]}{s\cdot M_d[\phi']},&\text{with
  }s=\frac{M_c[\alpha]}{M_d[\alpha]},\\
\intertext{so in particular we can merge $d$ into $c$ by first rescaling $d$:}
M_{cd}[\phi] &= \left\{\begin{aligned} &M_c[\phi] &\text{if }\phi \in c\\
&s\cdot M_d[\phi] &\text{otherwise}
\end{aligned}\right.\;.
\end{align*}
In general let $s^{cd}_\alpha = \frac{M_c[\alpha]}{M_d[\alpha]}$ for any $\alpha \in c
\cap d$ be the \emph{scale factor} of $c$ and $d$ w.r.t.\ $\alpha$.  Then in the case
that there are multiple plans in the intersection we 
are faced with multiple ways of performing a transitive closure, i.e.,
a set of scale factors.  These will normally be different from one
another, but, in the limit of data, assuming preferences and
feasibility constraints are actually independent of one another, every
scale factor between two clusters will be equal.  So, then, we take
the average:
\begin{align}
s^{cd} &:= \frac{1}{\size{c \cap d}}\cdot\sum_{\alpha\;\! \in\;\! c \;\!\cap\;\! d}
s^{cd}_\alpha,&\text{and,}\\
M_{cd}[\phi] &:= \left\{\begin{aligned} &M_c[\phi] &\text{if }\phi \in c\\
&s^{cd}\cdot M_d[\phi] &\text{otherwise}
\end{aligned}\right.\;.
\end{align}
In short, if all the assumptions are met, and enough data is given,
the described process will reproduce the correct prior distribution on
preferences.  Algorithm~\ref{alg:rescale} provides the remaining
details in pseudocode, and in the following we discuss these details
and the result of the rescaling process operating in the Travel domain.

\subsection{Rescaling}

\noindent{\bf Output.}  The result of rescaling is a set of clusters
of weighted plans, $C = \{ c_1, c_2, \ldots, c_n \}$.  Each cluster,
$c \in C$, consists of a set of plans with associated weights; we write $p
\in c$ for membership and $w_c(p)$ for the associated weight.

\begin{algorithm}
\label{alg:rescale} {
\small
\SetLine
\KwIn{Training records $\Phi$.}
\KwOut{Clusters $C$.}

initialize $C$ to empty \\
\ForAll{$(\phi,F) \in \Phi$} {\label{start_cluster}
    \eIf{$\exists c \in C$ such that $F \subseteq c$ or $F \supseteq c$}{
    \ForAll{$p \in F\setminus c$}{
      add $p$ to $c$ with $w_c(p) := \epsilon$
    }
    \eIf{$w_c(\phi) \geq 1$}{increment $w_c(\phi)$}{$w_c(\phi) := 1$}
    }{
      initialize $c$ to empty\\
      add $c$ to $C$\\
      \ForAll{$p \in F$}{
    add $p$ to $c$ with $w_c(p) := \epsilon$
      }
      $w_c(\phi) := 1$
    }
}
\label{end_cluster}

\While{$\exists c,d \in C$ such that $c \cap d \neq \emptyset$}{\label{start_merge}
    \label{start_scale}
    sum\_ratios $:= 0$\\
    \ForAll{$p \in c \cap d$} {
        sum\_ratios += $\frac{w_c(p)}{w_d(p)}$
    }
    scale := $\frac{\text{sum\_ratios}}{\size{\:\!c \;\cap\; d\:\!}}$\\
    \label{end_scale}

    \ForAll{$p \in d \setminus c$} {
      add $p$ to $c$ with $w_c(p) := w_d(p) \cdot \text{scale}$
    }
    remove $d$ from $C$
}
\label{end_merge}

\Return{$C$}
\label{end_combine}
}

%
%
%
%
%
%
%
%

\caption{Rescaling}

\end{algorithm}

\Ignore{

\subsection{Combining User Preferences Expressed Under Different Situations}
%
}

\smallskip\noindent{\bf Clustering.} First we collapse all of the input records
from the same or similar situations into single weighted clusters, with one count
going towards each instance of an observed plan participating in the
collapse.  For example, suppose we observe 3 instances of
\emph{Gobyplane} chosen in preference to \emph{Gobytrain} and 1
instance of the reverse in similar or identical situations.  Then we will end up with a cluster with weights 3 and 1 for \emph{Gobyplane} and
\emph{Gobytrain} respectively.  In other words
$w_c(p)$ is the number of times $p$ was chosen by the
user in the set of situations collapsing to $c$ (or $\epsilon$ if $p$ was never
chosen). This happens in lines~\ref{start_cluster}--\ref{end_cluster}
of Algorithm~\ref{alg:rescale}, which also defines `similar' (as set
inclusion).  Future work should consider more sophisticated clustering methods.

\smallskip\noindent{\bf Transitive Closure.} Next we make
indirect inferences between clusters; this happens by iteratively
merging clusters with non-empty intersections.  Consider two clusters,
$c$ and $d$, in the Travel domain.  Say $d$ contains \emph{Gobyplane} and
\emph{Gobytrain} with counts 3 and 1 respectively, and $c$ contains
\emph{Gobytrain} and \emph{Gobybike} with counts 5 and 1
respectively.  From this we infer that \emph{Gobyplane} would be
executed 15 times more frequently than \emph{Gobybike} in a situation
where all 3 plans (\emph{Gobyplane}, \emph{Gobytrain}, and
\emph{Gobybike}) are possible, since it is executed 3 times more
frequently than \emph{Gobytrain} which is in turn executed 5 times
more frequently than \emph{Gobybike}.  We represent this inference by
scaling one of the clusters so that the shared plan has the same
weight, and then take the union. In the example, supposing we merge $d$ into $c$, then we
scale $d$ so that $c \cap d = \{\text{Gobytrain}\}$ has the same
weight in both $c$ and $d$, i.e.\ we scale $d$ by $5 =
\frac{w_c(\text{Gobytrain})}{w_d(\text{Gobytrain})}$.
For pairs of clusters
with more than one shared plan we scale $d \setminus c$ by the average
of $\frac{w_c(\cdot)}{w_d(\cdot)}$ for
each plan in the intersection, but we leave the weights of $c \cap d$ as
in $c$ (one could consider several alternative strategies for plans in the intersection).\Ignore{
For pairs of clusters with more than one shared plan we treat the entire
intersection as one large ``meta-plan'' and scale $d\setminus c$ by $w_c(c \cap d) /
w_d(c \cap d) := (\sum_{p \in c \cap d} w_c(p)) / (\sum_{p \in c \cap
  d} w_d(p))$, and leave the weights of $c \cap d$ as in
$c$.}\Ignore{\footnote{A different possibility would be to average the shared
  plans: $w_{c'}(p) = (w_c(p) + \text{scale}*w_d(p)) / 2$.} }
Computing the scaling factor happens in
lines~\ref{start_scale}--\ref{end_scale} and the entire merging
process happens in lines~\ref{start_merge}--\ref{end_merge} of Algorithm~\ref{alg:rescale}.

\subsection{Learning}

We learn a set of pHTNs for $C$ by applying the base learner (with the
obvious generalization to weighted input) to each $c \in C$: 
\begin{align}
{\bf H} = \{ H_c = \text{EM}(\text{SH}(c)) \mid c \in C \}.
\end{align}
While the input clusters will be disjoint, the base learner may very well
generalize its input such that various pairs of plans become
comparable in multiple pHTNs within ${\bf H}$.  Any disagreement is
resolved by voting; recall that, given a pHTN $\calH$ and a plan $p$, we
can efficiently compute the most probable parse of $p$ by $\calH$ and
its (a priori) likelihood, say $\ell_\calH(p)$.  Given two plans $p$
and $q$ we let $\prec_\calH$ order $p$ and $q$ by $\ell_\calH(\cdot)$, i.e., $p \prec_\calH q
\iff \ell_\calH(p) < \ell_\calH(q)$;  if either is not parsable (or
tied) then they are incomparable by $\calH$.  Given a set of
pHTNs ${\bf H} = \{\calH_1, \calH_2, \ldots \}$, we take a simple
majority vote to decide $p \prec_{\bf H} q$ (ties are incomparable):
\begin{align}
p \prec_{\bf H} q  \iff \size{\{\calH \in {\bf H} \mid q
  \prec_\calH p \}} < \size{\{\calH \in {\bf H} \mid p \prec_\calH q \}}.
\end{align}
So, each pHTN votes, based on likelihood, for $p \prec q$ (meaning $p$ is
preferred to $q$), or $q \prec p$ ($q$ is preferred to $p$),
or abstains (the preference is unknown).  Summarizing, the input
$\Phi$ is 1) clustered and 2) transitively closed, producing clusters $C$, 
3) each of which is given to the base learner, resulting in a set
of pHTNs ${\bf H}$ modeling the user's preferences via the relation
$\prec_{\bf H}$.

\begin{figure}
 \centering
 \subfigure[]{
   \begin{minipage}[b]{0.5\columnwidth}
\label{results_rate}
     \includegraphics[width=1\textwidth]{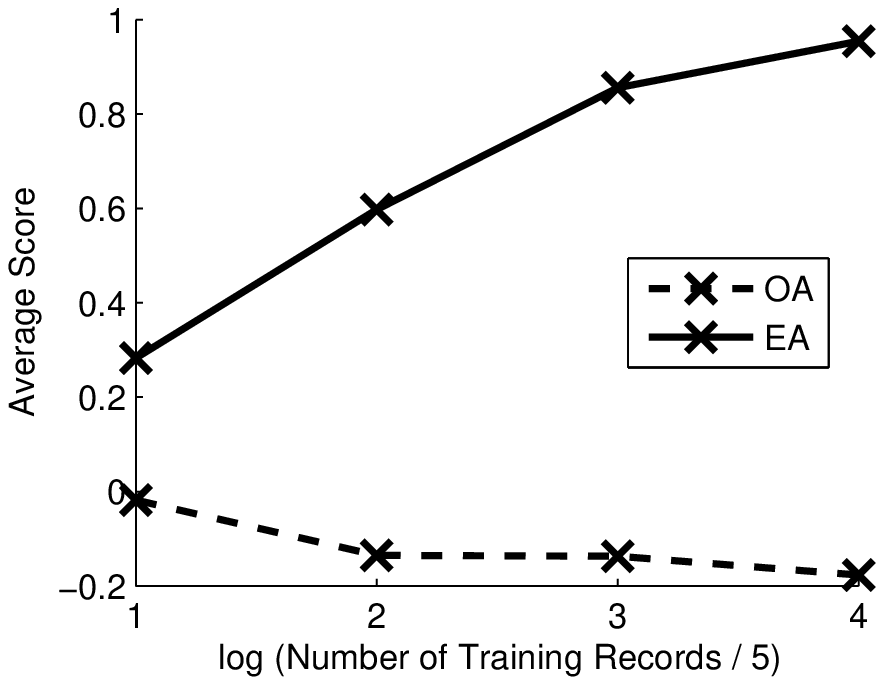}
   \end{minipage}
}%
\subfigure[]{
   \begin{minipage}[b]{0.5\columnwidth}
\label{results_size}
     \includegraphics[width=1\textwidth]{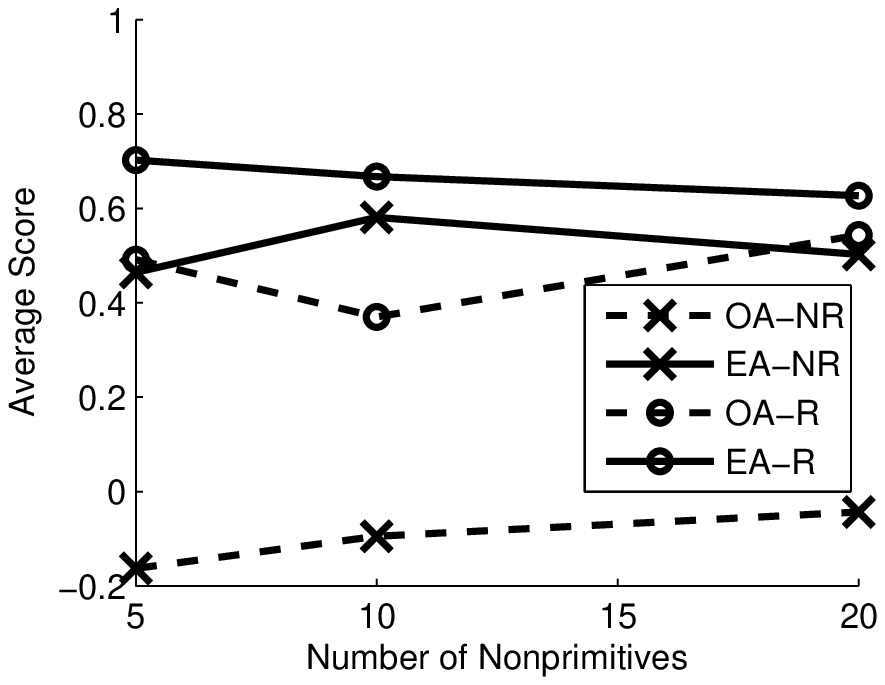}
    \end{minipage}
}%
\caption{Experimental results for random $\calH^*$.  ``EA'' is learning
 with rescaling and ``OA'' is learning without rescaling.  (a)
 Learning rate. (b) Size dependence: ``R'' for recursive $\calH^*$ and
 ``NR'' for non-recursive $\calH^*$.
}
\mvp
\end{figure}

\subsection{Evaluation}
%

\Ignore{In order to evaluate the ability of our preference learning
algorithm, we designed and carried out experiments in both synthetic
domains and benchmark domains.}

In this part we are primarily interested in evaluating the rescaling extension of
the learning technique, i.e., the ability to learn preferences despite
feasibility constraints.
\Ignore{While a direct user study is perhaps the obvious approach, user
studies are much too costly and anyways this would
involve cobbling together a number of other components (sensors, activity
recognizers, planner, human-appropriate interface, ...), and so in the
end credit/blame assignment would be difficult. For example it is
quite likely that the effectiveness of the interface would dominate
all other components in explaining user-reported satisfaction.  So w
}We
design a simple experiment to demonstrate that learning purely from
observations is easily confounded by constraints placed in the way of
user preferences, and that our rescaling technique is able to recover
preference knowledge despite obfuscation.\Ignore{does not learn what a user \emph{prefers} (but rather what
a user \emph{does}), and that our rescaling technique is .
evaluate rescaling with learning
against learning alone.  more controlled experiments for just the learner.}

\subsubsection{Setup}

{\bf Performance.} We again take an oracle-based experimental strategy, that is, we imagine a
user with a particular ideal pHTN, $\calH^*$, representing that user's
preferences, and then test the efficacy of the learner at recovering
knowledge of preferences based on observations of the imaginary
user.  More specifically we test the learner's performance in the
following game.  After training the
learner produces ${\bf H}_r$; to evaluate the effectiveness of ${\bf H}_r$ we
pick random plan pairs and ask both $\calH^*$ and ${\bf H}_r$ to pick the
preferred plan.\Ignore{We use two scoring systems ``+-'' and ``+''.}
There are three possibilities: ${\bf H}_r$ agrees with $\calH^*$ (+1 point), ${\bf H}_r$ disagrees
with $\calH^*$ (-1 point), and ${\bf H}_r$ declines to choose (0
points)\footnote{This gives rescaling a potentially
  significant advantage, as learning alone always chooses.  We also tested scoring ``no choice'' at -1 point; the
results did not (significantly) differ.\Ignore{ (in these
experiments).}}.\Ignore{ In the ``+-'' system the
learner receives +1, -1, and 0 points respectively; in the ``+''
system the learner receives +1, 0, and 0 points respectively.
}\Ignore{In both
systems the learner receives +1 point for a correct answer (agreeing
with $\calH^*$).  The ``+'' system does not penalize wrong answers (the
learner receives 0 points), whereas the ``+-'' system penalizes wrong
answers with -1 point.  Note the learner can decline to choose an
answer, and receives 0 points if so.
,
the learner, ${\bf H}_r$, is awarded +1 point if it picks the right
answer, and 0 points if it declines to choose among the pair.  In the
``+-'' system a penalty, of -1, is imposed for a wrong answer; in the
``+'' system no penalty is given for a wrong answer.
}

The distribution on testing plans is not uniform and will be described below.  The number
of plan pairs used for testing is scaled by the size of $\calH^*$; $100t$
pairs are generated, where $t$ is the number of non-primitives.  The final
performance for one instance of the game is the average number of
points earned per testing pair.\Ignore{, i.e., the accrued points divided by
$100t$.}\Ignore{Each plot further averages over 100 randomly selected
$\calH^*$.} Pure guessing, then, would get (in the long-term) $0$ performance.

\smallskip\noindent{\bf User.} As in the prior evaluation we evaluate
on 1) randomly generated pHTNs modeling possible users, and on 2) hand-crafted pHTNs
modeling our preferences in Logistics and Gold Miner.

\Ignore{
We carry out two kinds of experiments,
distinguished by the distribution on users.  In the first kind we randomly generate pHTNs, as in
the prior evaluation.  In the second kind we reuse the pHTNs (modeling
our preferences) created for Logistics and Gold Miner 
\Ignore{  These
experiments are further divided by generating either \emph{non-recursive} or
\emph{recursive} pHTNs.
In the second kind we design a particular pHTN by hand, modeling our own
preferences in Loa few benchmark domains.}
}


\smallskip\noindent{\bf Training Data.}
For both randomly generated and hand-crafted users we use the same
framework for generating training data.
We generate random problems by generating random solution sets in a
particular fashion, that is, \emph{we model feasibility constraints
using a particular random distribution on solution sets}.  We describe
this process in detail below, but note that the details are
unimportant (we did not try any alternatives to the choices given);
what matters is whether or not the process is a reasonable model of
the effect of feasibility constraints (insofar as they affect the
learning problem).

\Ignore{To generate random problems we
instead generate random solution sets (as there is no domain theory or
planner in the system (yet)).  That is, we \emph{model} feasibility
constraints using a particular random distribution on solution sets.\Ignore{
of solutions.}}

We begin by constructing a list of plans, $\calP$, from $100t$ samples
of $\calH^*$, removing duplicates (so $\size{\calP} \leq 100t$).
Due to duplicate removal, less preferred plans occur later than more
preferred plans (on average).  We reverse that order, and associate
$\calP$ with (a discrete approximation to) a power-law distribution.  Both
training and test plans are drawn from this distribution.
Then, for each training record $(\phi_i,F_i)$, we take a random
number\footnote{The number of samples taken is selected from
  $\size{\calP} \cdot \size{\cal N(0,1)} / 2$, subject to minimum 2 and
  maximum $|\calP|$, where ${\cal N}(.)$ is the normal distribution.  Larger solution sets model `easier' planning
  problems.} of samples
from $\calP$ as $F_i$.  \Ignore{So the least preferred plans of
$\calH^*$ are the most likely elements of $F_i$.} We sample the
observed plan, $\phi_i$, from $F_i$ by $\ell$, that is, the probability
of a particular choice $\phi_i = p$ is $\frac{\ell(p)}{\sum_{q\in F}
  \ell(q)}$.  \Ignore{So the most preferred, yet feasible (within
$F_i$), plan will be selected most often.}


\Ignore{Note that the learner does not have access
to $l(\cdot)$, as that information is strictly on the oracle side of the
testing framework, so that it really is necessary to explicitly pick $O(r)$
and inform the learner of the choice.
More importantly
n}

Note, then, that the random solution sets model the ``worst case'' of feasibility
constraints, in the sense that it is the least preferred plans that
are most often feasible --- much of the time the hypothetical user
will be forced to pick the least evil rather than the greatest good.

\smallskip\noindent{\bf Baseline.}  The baseline for our experiments
will be the original approach: the base learner without
rescaling.  That is, we take a single cluster, where the weight of
each plan is the number of times it is observed $w(\phi) = \size{\{
  i \mid \phi = \phi_i \}}$, and apply the base learner, obtaining a single
pHTN, ${\bf H}_b = \{ \calH \}$, and score it in the same manner that
the extended approach is scored by.

\subsubsection{Results: Random $\calH^*$}

\noindent{\bf Rate of Learning.} Figure~\ref{results_rate} presents
the results of a learning-rate experiment against randomly selected
$\calH^*$.  For these experiments the number of non-primitives is
fixed at 5 while the amount of training data is varied; we plot the
average performance, over 100 samples of $\calH^*$, at each training
set size.

We can see that with a large number of
training records, rescaling before learning is able to capture
nearly full user preferences, whereas learning alone performs
slightly worse than random chance. This is expected since without
rescaling the learning is attempting to reproduce its input
distribution, which was the distribution on observed plans --- and
``feasibility'' is inversely related to preference by construction.
That is, given the question ``Is $A$ preferred to $B$?'' the
learning alone approach instead answers the question ``Is $A$
executed more often than $B$?''. 

\Ignore{ We also tested the two
algorithms without imposing penalties for wrong answers (the ``+''
scoring system) in order  to ensure that the rescaling approach is
not benefiting unduly from the ability to decline choosing.  From
the results in Figure~\ref{results_rate} we can see that it is
slightly benefiting from this facility when there is little training
data \Ignore{(it performs less than 0.1 better than chance in the
``+'' system and more than 0.2 better than chance in the ``+-''
system)}, but this distinction disappears as the amount of training
increases.  This is expected, as with a large amount of training
data all records end up collapsing to a single cluster, and the
learner will no longer decline to choose when queried. With the
understanding that the scoring range is halved in the ``+'' system,
the learning alone approach behaves essentially the same, that is,
worse than chance. }

\smallskip\noindent{\bf Size Dependence.}  We
also tested the performance of the two approaches under varying number of
non-primitives (using $50t$ training records); the results are shown
in Figure \ref{results_size}.  For technical reasons, the base learner is much more effective at recovering
user preferences when these take the form of recursive schemas, so
there is less room for improvement.  Nonetheless the rescaling
approach improves upon learning alone in both experiments.

\subsubsection{Results: Hand-crafted $\calH^*$}

%

We re-use the same pHTNs encoding our preferences in Logistics and
Gold Miner from the first set of evaluations.  As
mentioned we use the same setup as in the random experiments, so it
continues to be the case that the distribution on random `solutions'
is biased against the encoded preferences.  Moreover, due to
the level of abstraction used (truncating to action names), as well as
the nature of the pHTNs and domains in question, the randomly
generated sets of alternatives, $F_i$, are in fact sets of solutions
to \emph{some} problem expressed in the normal fashion (i.e., as an
initial state and goal).

\smallskip
\noindent{\bf Logistics.} After training with 550 training records
($50t$, for 11 non-primitives) the baseline system scored only 0.342
(0 is the performance of random guessing) whereas rescaling before learning
performed significantly better with a score of 0.847 (0.153 away from
perfect performance).


\smallskip
\noindent{\bf Gold Miner:} After training with 600 examples ($50t$ for
12 non-primitives) learning alone scored a respectable 0.605, still,
rescaling before learning performed better with a score of 0.706.
Note that the greater recursion in Gold Miner, as compared to
Logistics, is both hurting and helping.  On the one hand the full
approach scores worse (0.706 vs. 0.847), on the other hand, the
baseline's performance is hugely improved (0.605 vs. 0.342).  As
discussed previously, the presence of recursion in the preference
model makes the learning problem much harder (since the space of
acceptable plans is then actually infinite), which continues to be a
reasonable explanation of the first effect (degrading performance).

The latter effect is more subtle.  The experimental setup, roughly
speaking, inverts the probability of selecting a plan, so that using a
recursive method many times in an observed plan is more likely than
using the same method only a few times.  Then the baseline approach is
attempting to learn a distribution skewed towards greater use of
recursion overall, and in particular, a distribution that prefers more recursion to
less recursion all else being equal.  However, there is no pHTN that
prefers more recursion to less recursion all else being equal; fewer
uses of a recursive method always increases the probability of a plan.  So the
baseline will fit an inappropriately large probability to any
recursive method, but, it will still make the correct decision between two plans differing only
in the depth of their recursion over that method. Naive Bayes
Classifiers exhibit a similar effect~\cite[Box 17.A]{probabilistic-graphical-models-book}.

\section{Discussion and Related Work}
\label{rel-work}





In the planning community, HTN planning has for a long time been given two
distinct and sometimes conflicting interpretations (c.f.
\cite{rao98}): it can be interpreted either in terms of domain
abstraction\footnote{Non-primitives are seen as abstract actions, mediating access to the
concrete actions.} or in terms of expressing complex (not first order
Markov) constraints on plans\footnote{Non-primitives are seen as
  standing for complex preferences (or even physical constraints).}.
The original HTN planners were motivated by the
former view (improving efficiency via abstraction). In this view, only top-down HTN planning
makes sense as the HTN is supposed to express effective search control.  Paradoxically, w.r.t.\ that motivation, the complexity of HTN
planning is substantially worse than planning with just primitive
actions~\cite{DBLP:journals/amai/ErolHN96}.  The latter view explains the
seeming paradox easily --- finding {\em a} solution should be easier, in general, than
finding one that also satisfies additional complex 
constraints. From this perspective both top-down \emph{and bottom-up}
approaches to HTN planning are appropriate (the former if one is
pessimistic concerning the satisfiability of the complex constraints,
and the latter if one is optimistic).  Indeed, this perspective lead
to the development of bottom-up approaches~\cite{barrett94}.

Despite this dichotomy, most prior work on learning HTN models
(e.g. \cite{ilghami02,langley06,yang07,hogg08}) has focused only on the
domain abstraction angle.
Typical approaches here require the structure of the reduction
schemas to be given as input, and focus on learning applicability
conditions for the non-primitives. In contrast, our work
focuses on learning HTNs as a way to capture user preferences, given
only successful plan traces.
The difference in focus also explains the difference in evaluation
techniques. While most previous HTN learning efforts are evaluated in terms of how close
the learned schemas and applicability conditions are, syntactically,
to the actual model, we evaluate in terms of how close the distribution
of plans generated by the learned model is to the distribution
generated by the actual model\Ignore{ (in the non-probabilistic setting we
would measure the similarity of the two sets of parsable plans)}.

An intriguing question is whether pHTNs learned to capture user
preferences can, in the long run, be over-loaded with domain
semantics. In particular, it would be interesting to combine the two
HTN learning strands by sending our learned pHTNs as input to the
method applicability condition learners. Presuming the user's
preferences are amenable, the applicability conditions thus learned
might then allow efficient top-down interpretation (of course, the
user's preferences could, in light of the complexity results for HTN
planning, be so antithetical to the nature of the domain that
efficient top-down interpretation is impossible).


%
%

As discussed in \cite{baier-aimag} there are other
representations for expressing user preferences, such as trajectory constraints expressed in linear
temporal logic. It will be
interesting to explore methods for learning preferences in those
representations too, and to see to what extent typical user preferences
are naturally expressible in (p)HTNs or such alternatives.



\section{Conclusion}
Despite significant interest in learning in the context of planning,
most prior work focused only on  learning domain physics or search
control.  In this paper, we expanded this scope by learning user
preferences concerning plans.  We developed a framework for learning
probabilistic HTNs from a set of example plans, drawing from the
literature on probabilistic grammar induction.  Assuming the input
distribution is in fact sampled from a pHTN, we demonstrated that the
approach finds a pHTN generating a similar distribution.  It is,
however, a stretch to imagine that we can sample directly from such a
distribution --- chiefly because observed behavior arises from a
complex interaction between preferences and physics.  

We demonstrate a technique overcoming the effect of such feasibility
constraints, by reasoning about the available alternatives to the observed user
behavior.  The technique is to rescale the
distribution to fit the assumptions of the baseline pHTN learner.  We
evaluate the approach, and demonstrate both 
that the original learner is easily confounded by constraints placed
upon the preference distribution, and that rescaling is effective at
reversing this effect.  We discuss several remaining important directions for
future work to address.  Of these, the most directly relevant technical
pursuit is learning parameterized pHTNs, or more generally,
learning conditional preferences.  Fully integrating an automated
planner with the learner, thereby using the learned knowledge, and
running (costly...) user studies are also very important pursuits.
In the end, we describe an effective approach to automatically
learning a model of a user's preferences from observations of only
their behavior.

\medskip
\noindent
{\bf Acknowledgments:} Kambhampati's research is supported in part by ONR grants
N00014-09-1-0017 and N00014-07-1-1049, and the DARPA Integrated
Learning Program (through a sub-contract from Lockheed Martin).




\bibliographystyle{elsarticle-num} 
\bibliography{icaps09}

\Ignore{

}
\end{document}